# Semi-supervised multiscale dual-encoding method for faulty traffic data detection

Yongcan Huang[1] and Jidong J. Yang[1]

[1]Smart Mobility and Infrastructure Laboratory, College of Engineering, University of Georgia, Athens GA, USA

**Abstract**

Inspired by the recent success of deep learning in multiscale information encoding, we introduce a variational autoencoder (VAE) based semi-supervised method for detection of faulty traffic data, which is cast as a classification problem. Continuous wavelet transform (CWT) is applied to the time series of traffic volume data to obtain rich features embodied in time-frequency representation, followed by a twin of VAE models to separately encode normal data and faulty data. The resulting multiscale dual encodings are concatenated and fed to an attention-based classifier, consisting of a self-attention module and a multilayer perceptron. For comparison, the proposed architecture is evaluated against five different encoding schemes, including (1) VAE with only normal data encoding, (2) VAE with only faulty data encoding, (3) VAE with both normal and faulty data encodings, but without attention module in the classifier, (4) siamese encoding, and (5) cross-vision transformer (CViT) encoding. The first four encoding schemes adopted the same convolutional neural network (CNN) architecture while the fifth encoding scheme follows the transformer architecture of CViT. Our experiments show that the proposed architecture with the dual encoding scheme, coupled with attention module, outperforms other encoding schemes and results in classification accuracy of 96.4%, precision of 95.5%, and recall of 97.7%.

## 1. Introduction

State departments of transportation in the U.S. typically collect traffic data using permanent continuous count stations (CCS), equipped with inductive loop sensors that routinely require checking and calibration to ensure quality data being collected [1]. It is recommended that accuracy, completeness, validity, timeliness, coverage, and accessibility are used as the major data quality measures [2]. However, inherent defects, disrepair, communication failure, environmental effects, among others, inevitably yield faulty CCS data. Thus, it is critical to have quality control of CCS data to support various transportation planning/engineering practices and decision-making that demands high-quality traffic data. The existing statewide quality control processes for CCS data feature rule-based mechanisms that either reject or flag the raw data based on established quality control rules. For ambiguous situations, further review is required by an analyst.

The faulty data detection has been extensively studied by typically following one of the three major approaches: (1) statistical analysis, (2) classical signal processing, and (3) artificial intelligence (AI) based methods [3]. Detecting faulty data by statistical analysis of historical data usually relies on some defined confidence intervals for identification of outliers. Principal components analysis (PCA) and its variants have been employed as well. For example, to optimize fault detection and diagnosis in nonlinear dynamical systems, Bounoua and Bakdi (2021) developed fractal-based dynamic kernel PCA, which overcomes the shortcomings of other variants of PCA [4].

In signal processing-based fault detection, wavelet transform, which allows time-series data to be portrayed in time and frequency domains, has been widely used. By exerting scaling and translation it allows comprehensive analysis of faulty signals in multiple scales and resolutions [5]. Continuous wavelets transform (CWT) has demonstrated success in extracting regular and irregular features (e.g., seasonal, trend, surge) [6]. CWT can convert time-series data to time-frequency image representation in desirable resolution, which has advantages over the Fourier transform techniques for fault detection [7–9].

For the AI-based methods, signal processing is usually applied first as a pre-processing step. For instance, to detect and classify faults in the shunt compensated static synchronous compensator transmission line, Aker et al. [10] proposed a probabilistic neural network-based Naïve Bayes approach where discrete wavelet transform assists with extracting features that are later used to train the classifiers to categorize faulty signals occurring in the system. In another study, Boquet et al. [11] introduced a variational autoencoder (VAE) model for prediction-based fault detection that learns how traffic data is generated in an unsupervised way. The traffic data is first projected to the low-dimensional latent space before feeding to the prediction model. More recently, Morris et al. [12] presented a VAE-based approach, where a joint latent space was constructed using two VAEs trained on image representations from two distinct signal processing techniques: CWT and recurrent plot (RP). Anomalous data is detected if the learned representation falls farther away from the approximated manifold in the latent space. Additionally, two independent data sources are cross-checked to verify an anomaly.

With the focus on the quality control of statewide traffic data, many state departments of transportation have adopted their own automatic review systems that use pre-defined quality control rules for data screening. However, the quality control rules rely on certain thresholds, which are subjective and insensitive to variation inherited in traffic data. Thus, they are not considered robust enough to detect faulty data and often lead to false-positive diagnosis. For example, the faulty data arising from minor sensor malfunction may not be detected, while an abnormal data pattern caused by an extraneous event (e.g., a traffic accident) is not faulty in traffic data itself but may be flagged as faulty data.

The simplest form of data quality control can be cast as a binary classification task, where faulty data are separated from normal data. However, the main challenge lies in proper construction of feature space for the classification task. In this respect, deep convolution neural networks have been commonly employed as backbones to extract multiscale features, which are then utilized for the classification task.

Motivated by both signal processing and AI-based fault detection approaches, a novel semi-supervised dual-encoding architecture is introduced for detection of faulty traffic data in this paper. First, time-series traffic volume data is transformed by CWT to generate time-frequency image representations. Two identical VAEs are then trained to encode normal and faulty CWT images in multiple scales separately. These multiscale encodings are subsequently utilized to classify whether the data is normal or faulty. By leveraging positive and/or negative multiscale encodings from the VAE encoders, four different VAE-based models are designed and evaluated. Two additional encoder architectures, including a siamese network (SN) encoder and a cross vision transformer (CViT) encoder, are used and evaluated as well for comparison purposes. The CCS data used in this study is obtained from the Georgia Department of Transportation.

The remainder of this paper is organized as follows: Section 2 reviews two key techniques, CWT and deep learning-based feature extraction adopted in our proposed method and their applications in fault detection with time-series data. Section 3 discusses our proposed method. The datasets used in this study are described in Section 4, followed by experiments in Section 5. The conclusions are given in Section 6 with discussions on future research directions.

## 2. CWT and deep learning-based feature extraction

This review section focuses on CWT and deep learning-based feature extraction and their applications in fault detection with time-series data.

*2.1. CWT in fault detection*

CWT is well known for decomposing time-frequency information, particularly, constructive to obtain salient features from dynamic time series data. CWT-based traffic data transformation has been shown to reveal unobvious patterns of traffic data in an efficient way [13].

Traditionally, CWT has been used to capture local changes, which are noisy and aperiodic. For example, Zheng et al. [14] demonstrated the utility of wavelet transform in analyzing important features associated with abnormal traffic conditions, such as bottleneck effects and traffic oscillation arising from congestion. The case study of three different scenarios of vehicle trajectories showed that the origins of deceleration waves could be detected by wavelet-based energies of a single vehicle, and the detected origins help to pinpoint possible causes. In another study, Jiang et al. [7] developed a two-stage fault detection method for anomalous network traffic. In their methodology, CWT was applied to decompose the incoming signals into multiple continuous scales, followed by principal component analysis to extract the features of anomalous network traffic. Then, a new mapping function is constructed to detect the abnormal traffic.

Recently, CWT coupled with deep learning techniques offers a new approach for fault detection with time-series data. König et al. [15] proposed a deep learning-based method for anomaly detection and diagnosis on acoustic emission signals. With the acoustic emission signals being converted to CWT images, an autoencoder network was developed for anomaly detection in the latent space and GoogLeNet [16] was adapted to the anomaly classification task. In another study, Jalayer et al. [17] developed a comprehensive deep learning-based fault detection and diagnosis model for rotating machinery by channeling up fast Fourier transform, CWT, and statistical features of raw signals. A convolutional long short-term memory was employed to classify the multi-channel input.

Based on the review of previous studies, CWT has been commonly used for processing time-series data. Generally, the methodology of converting signals into CWT images representations and further processing these image representations by deep-learning-based methods to encode multiscale features has shown a great potential and improved performance in fault detection of time-series data.

*2.2. Deep-learning based classification with time-series data*

Deep learning has been largely fueled by convolutional neural networks (CNNs) since AlexNet [18]. Nevertheless, its impacts and applications have gone far beyond the field of computer vision. This section reviews the applications of CNNs for classification of time series data.

Wang and Oates [19] encoded time-series data into the Gramian angular summation/difference field (GASF/GADF) and the Markov transition field (MTF) images and applied deep tiled convolutional neural networks for classification with 20 standard datasets. Hatami et al. [20] used RP to portray time-series to images and then designed a CNN classifier with 2 hidden layers to solve a time-series classification problem. It was compared with traditional machine learning frameworks as well as using other time-series images (e.g., GAF-MTF images). Pelletier et al. [21] studied temporal CNN for classification of land cover using sentinel-2 satellite image time series data. The proposed temporal CNN outperformed traditional random forest and recurrent neural networks in terms of overall accuracy. Yang et al. [22] exploited three time series imaging methods, GASF, GADF and MTF, and concatenated them into different channels as input to a CNN classifier. The study aimed to evaluate the impacts of different imaging methods, the sequence of image concatenation, and the complexity of CNN on classification accuracy. The results showed that the selection of imaging methods and the sequence of concatenation did not have significant impacts on the prediction outcome, and a simple CNN appeared to be sufficient for the classification task as compared to complex CNNs, such as VGG net [23].

More recently, Shi et al. [24] presented an image-based fault detection pipeline for nuclear power plants. The study casts the fault detection problem into a supervised image classification task. Four distinguished time series imaging methods, including GASF, GADF, MTF and un-thresholded recurrence plot (UTRP), were employed to transform the time-series data into two-dimensional image representations. The CNN-based, transformer-based, and MLP-based deep learning architectures were evaluated for feature extraction and classification. The experimental results showed that UTRP achieves excellent performance and work well across different deep-learning architectures. Specifically, the EfficientNet-B0, one of the CNN-based models evaluated, outperforms the other deep learning models in terms of the trade-off between speed and accuracy.

Based on our literature review, imaging time series data as a pre-processing step, followed by deep-learning based models (e.g., CNN and Transformer) is an effective approach for faulty data detection. In this study, self-supervised learning methods, such as CNN-based VAE, are applied to encode normal traffic data (referred to as positive encoding) and faulty traffic data (referred to as negative encoding) separately. Our study demonstrated that the pooling of complementary positive and negative features through an attention mechanism leads to improved fault detection performance.

## 3. VAE-based dual-encoding method for faulty traffic data detection

The conceptual framework of the proposed method is illustrated in Figure 1. First, the time series traffic data is partitioned to three datasets: positive dataset, negative dataset, and mixed dataset. The CWT images for the positive dataset and the negative dataset are used to train twin VAEs, referred to as $VAE^{(P)}$ and $VAE^{(N)}$ in this paper. The resulting multiscale dual encodings from $VAE^{(P)}$ and $VAE^{(N)}$ are concatenated and fed to a classifier, which consists of a self-attention module and an MLP. The mixed dataset, including both positive and negative examples, is used to train the classifier in a supervised fashion while the VAE encoders are kept frozen during the classifier training.

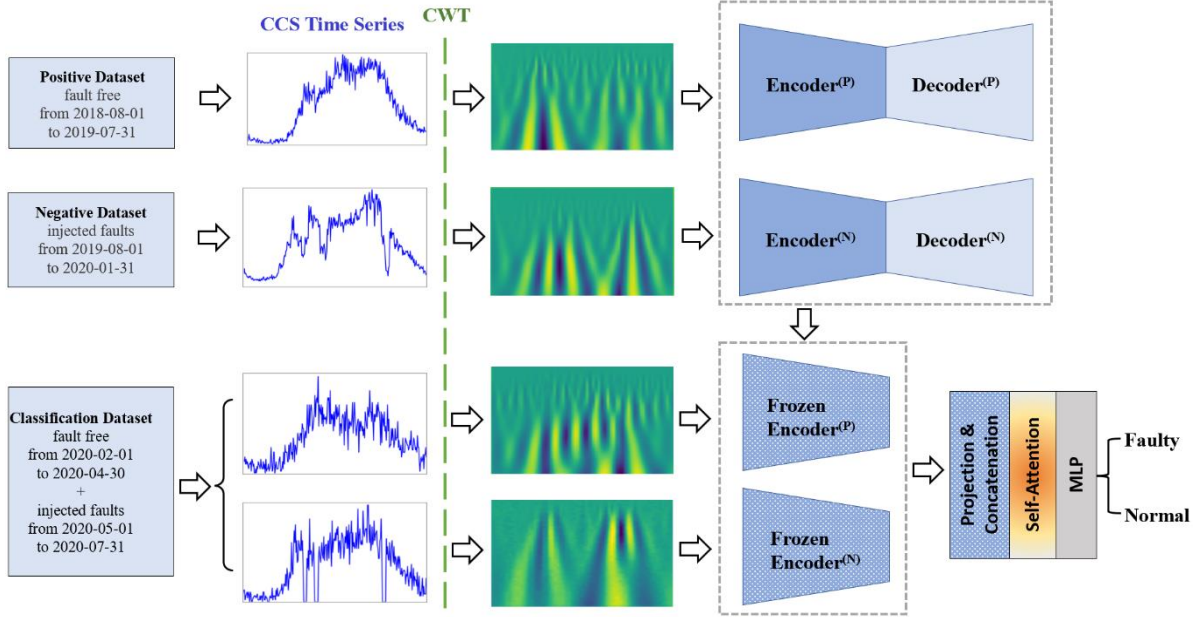

**Figure 1.** Illustration of the proposed fault detection methodology; the pretrained twin encoders are frozen and used to obtain multiscale dual encodings for training the classifier.

*3.1. Continuous wavelet transforms*

Wavelets are formed by convoluting scaled and translated versions of a chosen mother wavelet over time series data. In this context, scaling refers to stretching or shrinking the wavelet in time, where stretching the wavelet helps to capture slow changes in time series data, while shrinking it manifests the abrupt changes. Shifting refers to the position of the wavelet imposed on the signal. As a result, CWT produces a time-frequency image that visually captures changes in various scales, inducing a powerful representation for time-series data. The CWT function $f(t)$ is expressed in Equation 1.

$$C(a,b;f(t),\psi(t)) = \int_{-\infty}^{\infty} f(t)\frac{1}{a}\psi*(\frac{t-b}{a})dt \tag{1}$$

where, *a* represents the scale, and *b* indicates the position. The symbol * denotes the complex conjugate. $\psi$ represents the mother wavelet function. Morlet, one of the widely applied mother wavelet functions, is adopted in our study [25]. Figure 2 depicts an example of the Morlet application to the CCS data.

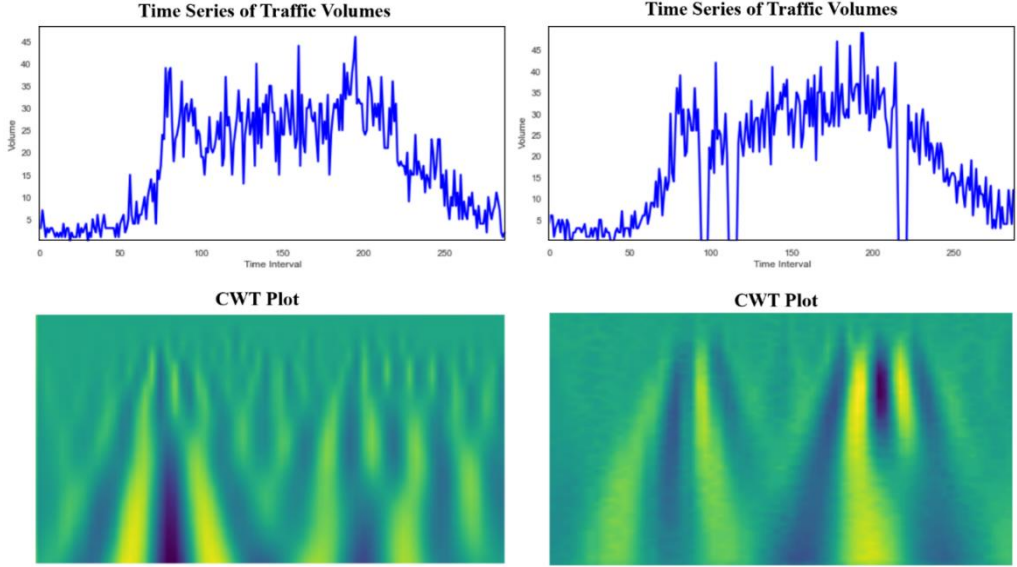

**Figure 2.** Sample wavelet transformations on CCS data; left: normal data, right: faulty data.

*3.2. VAE-based architectures for image classification*

CNNs have been widely used as feature extractors in image classification since the breakthrough demonstrated by AlexNet [18]. Given the prowess of CNNs in hierarchical features extraction, the deep convolutional VAE architecture is well suited to extract multiscale representations of CWT images.

Our proposed classifier leverages the multiscale positive and negative encodings from the pretrained $VAE^{(P)}$ and $VAE^{(N)}$ through an attention mechanism for the downstream classification task. This dual encoding scheme was inspired by [26], where a frozen dictionary learning method is proposed to learn a dictionary-based sparse representation of normal data and anomalous data in sequence. The portion of dictionary for normal data is learned first and then frozen while learning the portion of dictionary for anomalous data. Different from [26], we independently learn VAE encodings for the positive dataset and the negative dataset in multiple scales and then leverage attention mechanism to selectively aggregate features for the classification task. The remainder of this section introduces background on VAE and the self-attention mechanism, followed by our proposed VAE-based classifier.

3.2.1. Variational autoencoders

VAEs encode inputs through regularized reconstruction via an encoder-decoder paradigm [27], as depicted in Figure 3. The encoder transforms the input ($x$) in a low-dimensional latent space ($z$), while the decoder operates on $z$ to reconstruct the input ($\hat{x}$). To allow the underlying data structure (manifolds) to emerge in the low-dimensional latent space, the information content that could be encoded is constrained by injecting noises in $z$ via Gaussian sampling. Since directly sampling from $z$ is not a differentiable operation, a reparameterization trick is employed [27]. The loss function of the originally proposed VAE consists of two terms: (1) reconstruction loss ($l_r$), and (2) regularization ($l_{KL}$), where $\alpha$ is the weight given to the regularization term.

$$l(x, \hat{x}) = l_r + \alpha l_{KL}(z, N(0, I)) \qquad (2)$$

To learn the respective normal and faulty features, the twin VAEs are trained, one with the positive dataset and one with the negative dataset. The inputs are CWT images ($1 \times 64 \times 64$). The encoder of each VAE consists of four convolution blocks, where each block includes two 2D convolutional layers plus a max pooling, resulting in a 2048-long vector after flattening. Each convolutional layer is followed by batch norm and rectified linear unit (ReLU) activation. A fully-connected layer is used to further reduce the dimension from 2048 to 64. The decoder applies four transposed convolutional blocks to reconstruct the CWT image, where the first three blocks consist of two transposed convolutional layers and the fourth has an extra 2D convolutional layer. Samples of original CWT images and the reconstructed ones are shown in Figure 4. The two pretrained encoders, Encoder$^{(P)}$ and Encoder$^{(N)}$, are subsequently employed to extract multiscale features for the classification task.

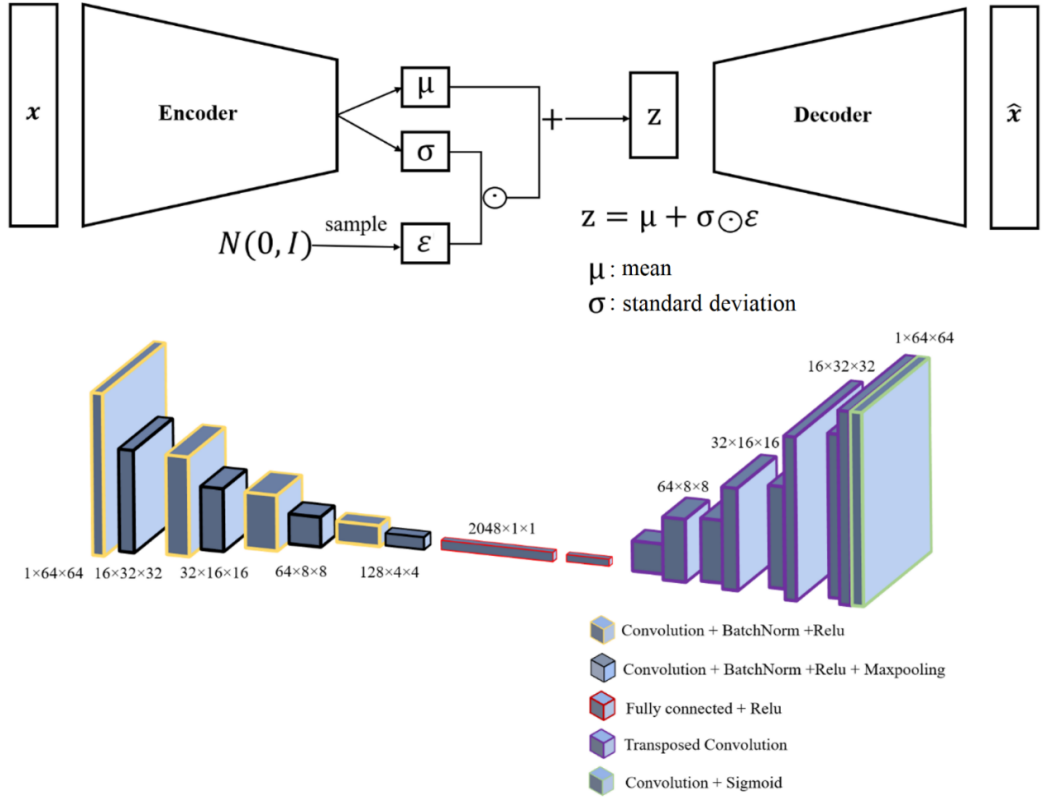

**Figure 3.** The top plot indicates the VAE framework. The bottom plot illustrates the proposed VAE architecture.

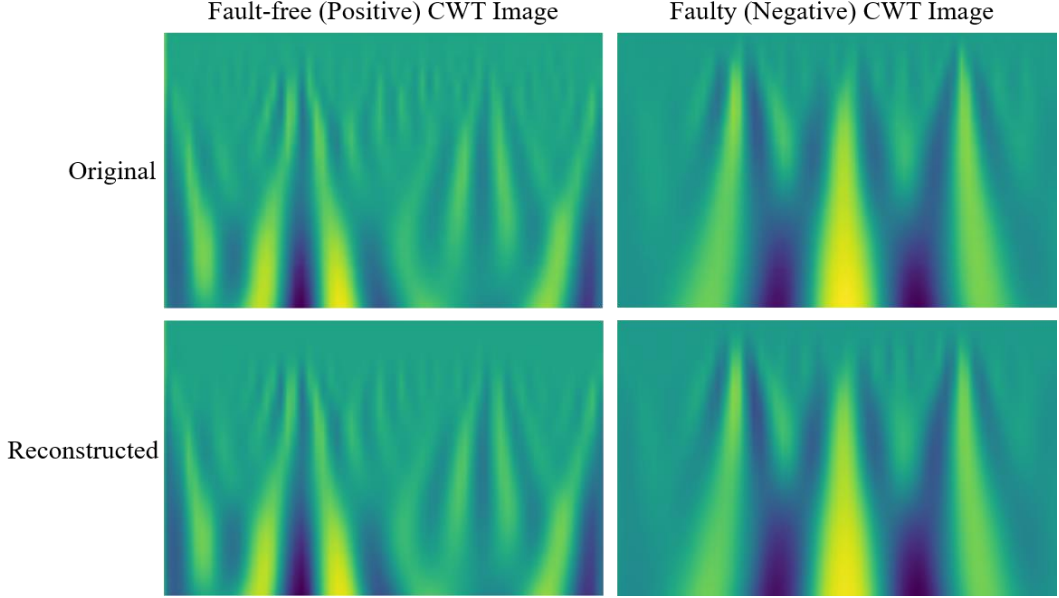

**Figure 4.** Samples of CWT images and corresponding reconstruction.

3.2.2. Attention-based classifier

The attention mechanism has become extremely popular, largely due to the success of transformer [28], and has been successfully applied for image classification task [29]. Inspired by human visual attention mechanism, an attention module pools information by paying due attention to different inputs through a weighting mechanism. Considering that not all features coming from the pretrained twin VAE encoders contribute at the same level to the classification task, a self-attention module is employed for selective feature pooling, as shown in Figure 5.

The features retrieved from each block of the pretrained Encoder[(P)] and Encoder[(N)] are firstly projected to a 1024-long vector. The resulting common-length vectors are then concatenated to obtain a matrix of size $8 \times 1024$. The attention mechanism is applied among the eight vectors, which represent the positive or negative features at different scales. Linear transformation is used to obtain the query ($Q$), key ($K$), and value ($V$) matrices of the same size. The scaled dot-product attention [28] is computed by Equation (3).

$$Attention(Q, K, V) = softmax\left(\frac{QK^T}{\sqrt{d_k}}\right)V \qquad (3)$$

where $d_k$ denotes the dimension of $K$ and $Q$

The output from the attention module is flattened and fed to a multilayer perceptron (MLP), which consists of two fully connected layers, followed by ReLU activation and a dropout layer ($p = 0.5$).

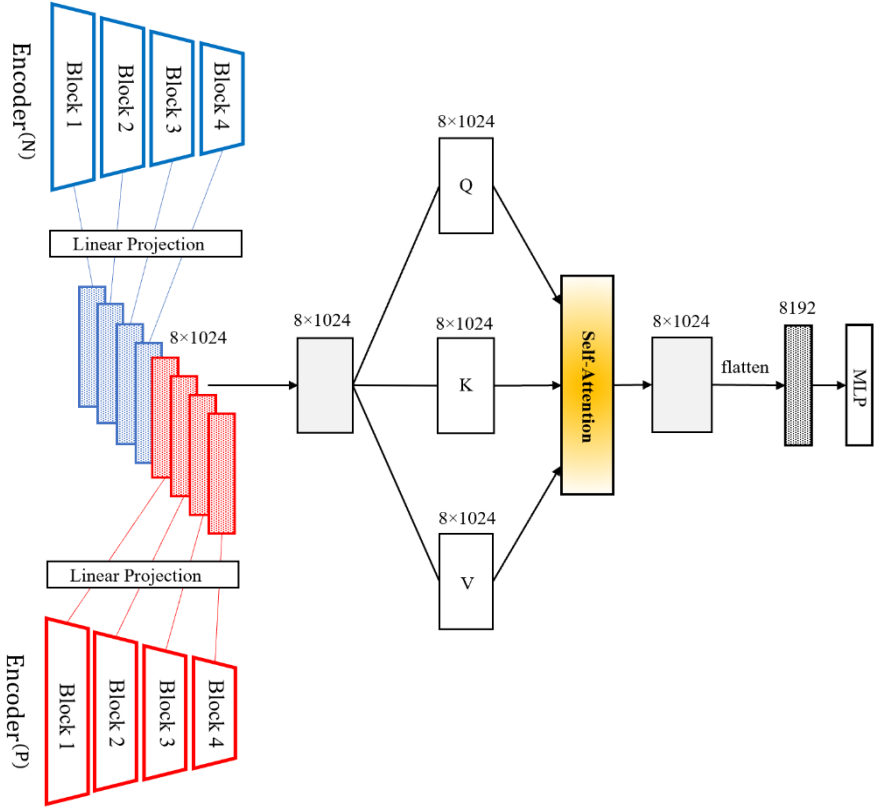

**Figure 5.** Architecture of the attention-based classifier.

## 4. Time series traffic datasets

Time series traffic volumes in 5-minute intervals are obtained from 254 active CCS sites across the state of Georgia over a two-year period (August 2018 - August 2020). The geospatial locations of CCS sites are depicted in Figure 6. Three distinct datasets, including positive dataset, negative dataset, and mixed dataset, are described below and summarized in Table 1.

*Positive dataset*. The positive dataset contains 81,443 normal daily time sequences of 5-minute traffic volumes, collected at 254 CCS sites from August 1st, 2018 to July 31st, 2019.

*Negative dataset*. Since it is challenging to naturally obtain faulty data on a large scale, this negative dataset was created by injecting artificial fault signals to simulate the actual faulty signals observed. Based on our observations [12], three major types of faulty signals present in the data, referred to as (1) point fault, (2) block fault, and (3) sensor nonresponsive fault. To be consistent with the observed natural data anomalies, three methods were used to create faulty signals: (1) decreasing volumes of five randomly selected time intervals by 40%, (2) randomly selecting sequential intervals of length $k$ ($k$ = 5 or 10) within a day and decreasing the volumes by 40%, and (3) suppressing random segments of length $k$ ($k$ = 5 or 10) to zero. Figure 7 demonstrates the faulty signals, where the thicker gray lines trace the normal data trends and the deviation points of the red lines from the gray lines indicate the faulty signals. A total of 40,864 days of normal traffic data from August 1, 2019 to January 31, 2020 were used to create this negative dataset.

*Mixed dataset*. This dataset is used to train the attention-based classifier. It contains 20,015 days of 5-minute traffic volume data from February 1, 2020 to April 30, 2020 and 20,599 days of "faulty" data, which are generated in the same manner by injecting faulty signals to the normal traffic volume data collected from May 1, 2020 to July 31, 2020.

It should be noted that the original normal data used to generate the negative data was excluded from this study to avoid information leakage. Table 1 summarizes the partition of CCS data for generating the three datasets.

**Table 1.** Summary of datasets.

| Dataset | Usage | Time window | Size |
| --- | --- | --- | --- |
| Positive dataset | Train/evaluate the $VAE^{(P)}$ | August 1, 2018- July 31, 2019 | 81,443 days |
| Negative dataset | Train/evaluate the $VAE^{(N)}$ | August 1, 2019- January 31, 2020 | 40,864 days |
| Mixed dataset | Train/evaluate the classifier | February 1, 2020- July 31, 2020 | 40,614 (20,015 normal days and 20,599 faulty days) |

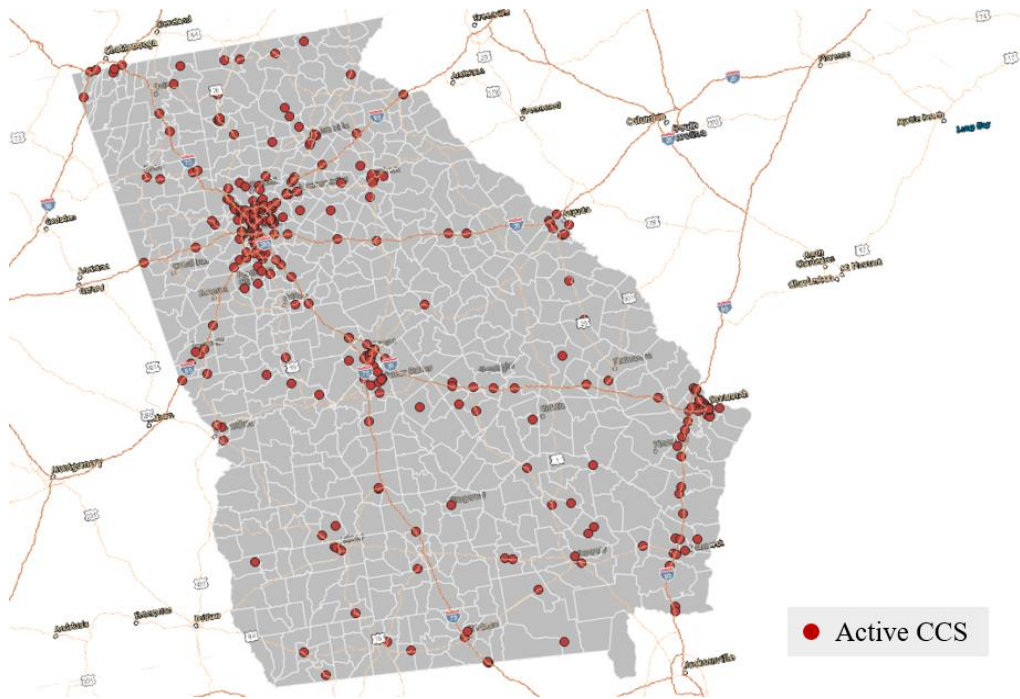

**Figure 6.** Locations of active CCS in Georgia, USA.

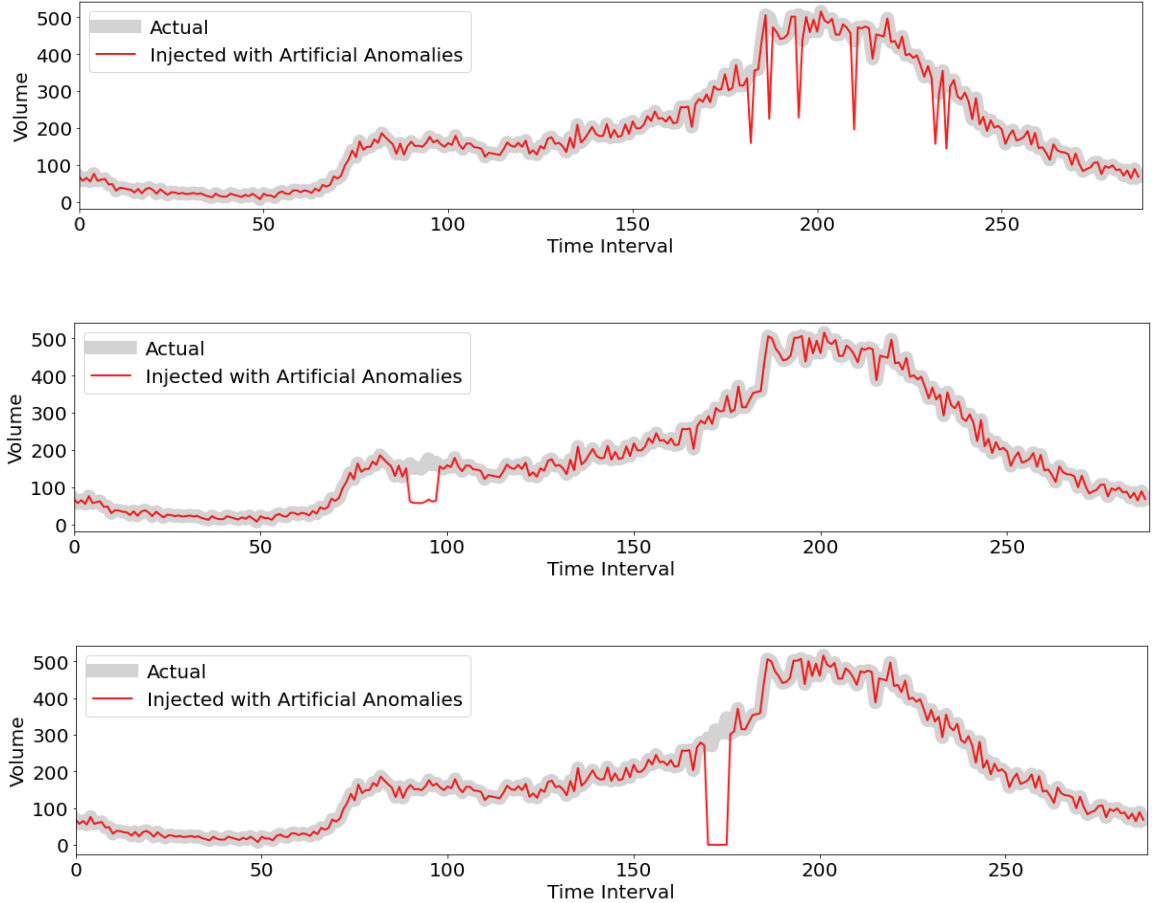

**Figure 7.** Visualization of three types of faulty signals. The top plot indicates point fault, which abruptly drops to an abnormal value; the middle plot reflects block fault, which has an extended period of low-level values; and the bottom plot demonstrates sensor nonresponsive fault, which displays a short window of zero [12].

## 5. Experiments

There are two stages of training for all VAE-based models. In the first stage, multiscale positive and negative encodings are obtained by training the twin VAEs: one using the positive dataset and the other using the negative dataset. The resulting VAE encoders serve as feature extractors. In the second stage, the VAE encoders are frozen, and the corresponding classifiers are trained using the mixed dataset. The Siamese network (SN) model [30] also involves two stages of training. The base network is first trained using contrastive loss and then frozen and used as the feature extractor for the classifier training in the second stage. The CViT [31] model is trained end to end.

The PyWavelets package [32] is used to convert daily time-series traffic volume data to CWT images ($1 \times 64 \times 64$). The VAE-based models and SN model are trained using Adam [33] with a learning rate of 0.001, a batch size of 32, and early stopping, where the minimum validation loss is sought. For training of the classifier, cross-entropy loss is used. The dataset is split into three sub-datasets: training set (60%), validation set (20%) and testing set (20%). All classifiers are trained for 50 epochs using Adam with a learning rate 0.00001 and a batch size of 32. The classifiers' performance on the test dataset is summarized in Table 2. The receiver operating characteristic (ROC) curves are plotted in Figure 8 with reported area under the ROC curve (AUC).

As shown in Table 2 and Figure 8, all models achieved good classification results. The proposed $VAE^{(PNS)}$, which leverages the dual encodings of positive and negative features through a self-attention mechanism, outperforms the other VAE-based models. This attests to the effectiveness of the attention mechanism for feature assembling in the classification task. Interestingly, the $VAE^{(PNS)}$ outperforms the SN and CViT as well.

By comparing $VAE^{(P)}$ and $VAE^{(N)}$, the slight improvement in accuracy from 89.0% to 91.5% is likely due to the fact that $VAE^{(P)}$ simply learned features from the normal dataset while $VAE^{(N)}$ learned features from the faulty dataset, which contains negative features as well as partial positive features, leading to the better classification result. The performance is improved when both positive and negative features are considered for classification task, evidenced by the accuracy of 93.1% for $VAE^{(PN)}$. The addition of the self-attention layer ($VAE^{(PNS)}$) further boost the accuracy to 96.4%, demonstrating the effectiveness of the attention module in feature assembling. In comparison, the other two popular network architectures (i.e., SN and CViT) have slightly inferior performance than the $VAE^{(PNS)}$.

**Table 2.** Model performance evaluation.

| Model | Backbone | Classifier | Precision | Recall | F1 score | Accuracy |
|---|---|---|---|---|---|---|
| $VAE^{(P)}$ | $Encoder^{(P)}$ | MLP | 86.3 % | 94.2 % | 90.1 % | 89.0 % |
| $VAE^{(N)}$ | $Encoder^{(N)}$ | | 91.5 % | 91.2 % | 91.3 % | 91.5 % |
| $VAE^{(PN)}$ | [$Encoder^{(P)}$, $Encoder^{(N)}$] | | 93.5 % | 93.3 % | 93.4 % | 93.1 % |
| $VAE^{(PNS)}$ | [$Encoder^{(P)}$, $Encoder^{(N)}$] | Self-Attention + MLP | 95.5 % | 97.7 % | 96.6 % | 96.4 % |
| SN | Siamese Encoder * | MLP | 94.5 % | 94.5 % | 94.5 % | 94.5 % |
| CViT [34] | Cross_ViT ** | | 94.4 % | 94.1 % | 94.3 % | 94.4 % |

Notes:

The VAE and SN backbones were frozen while the corresponding classifiers are trained.

* The Siamese Encoder shares the same architecture as the VAE Encoders

** ViT design: vector dimension = 1024, number of transformer blocks = 6, number of heads in multi-head attention layer = 16, dimension of the feedforward layer = 2048, dropout = 0.1, embedding dropout = 0.1

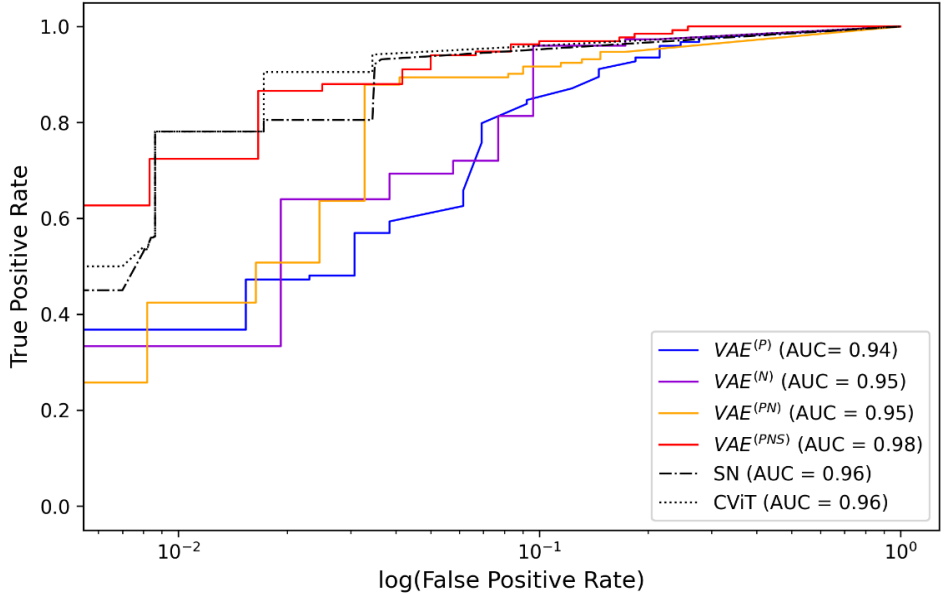

**Figure 8.** ROC curves and AUCs of all tested models. The false positive rate is plotted on logarithmic scale for better visualization.

## 6. Conclusions and future directions

This study focuses on the quality control of traffic data collected in time sequence and casts the detection of faulty traffic data as a classification problem. By leveraging time-series data transformation techniques and modern deep learning architectures for multiscale feature encodings, our proposed model is characterized by three major components: (1) CWT transformation of time-series traffic data, (2) a twin of pretrained VAE encoders for dual encodings of positive and negative multiscale features from CWT images, and (3) an attention-based classifier. Our experiments with purposely designed architectures demonstrated that harnessing both positive and negative features embodied in multiple scales through attention mechanism leads to better performance. This can be interpreted by the power of attention mechanism in pooling complementary contributions of positive and negative features for the classification task.

One drawback of the proposed model is the need for pretraining VAEs. To continuously learn and adapt to emerging features over time, meta-learning frameworks or appropriate finetuning procedures should be explored. Additionally, the current study is centered on one-dimensional time-series data. Further studies are needed to extend the framework to high-dimensional correlated time series data for broader applications. For example, this study focuses on individual traffic count stations while the flow patterns of geographically close stations are inherently correlated and constrained by the network topology. Therefore, it would be more effective to analyze clusters of stations for faulty data detection.

**Acknowledgments**

The work presented in this paper is part of a research project (RP 20-29) sponsored by the Georgia Department of Transportation. The contents of this paper reflect the views of the authors, who are solely responsible for the facts and accuracy of the data, opinions, and conclusions presented herein. The contents may not reflect the views of the funding agency or other individuals. The authors would like to acknowledge the financial support provided by the Georgia Department of Transportation for this study.